%% file: main.tex
\documentclass[10pt,twocolumn,letterpaper]{article}

\usepackage{cvpr}              
\usepackage{amsmath}
\usepackage{makecell}
\input{preamble}
\definecolor{cvprblue}{rgb}{0.21,0.49,0.74}
\usepackage[pagebackref,breaklinks,colorlinks,allcolors=cvprblue]{hyperref}

\def\paperID{45597} 
\def\confName{CVPR}
\def\confYear{2026}

\title{OptiMVMap: Offline Vectorized Map Construction via Optimal Multi-vehicle Perspectives}

\author{
    Zedong Dan$^{1,2}$\textsuperscript{\dag}
    ,
    Zijie Wang$^{1,4}$\textsuperscript{\dag}
   ,
    Wei Zhang$^{3}$,
    Xiangru Lin$^{1}$,
    Weiming Zhang$^{3}$,
    Xiao Tan$^{3}$,\\
    Jingdong Wang$^{3}$,
    Liang Lin$^{1,5}$,
    Guanbin Li$^{1,4,5}$\\[1ex]
    $^1$
    Sun Yat-sen University \qquad $^2$Zhongguancun Academy \qquad 
    $^3$Baidu Inc. \\ $^4$Shenzhen Loop Area Institute \qquad 
    $^5$Guangdong Key Laboratory of Big Data Analysis and Processing \\
    \small
    \texttt{\{danzd, wangzj75\}@mail2.sysu.edu.cn, liguanbin@mail.sysu.edu.cn}\\
}

\begin{document}
\maketitle
\def\thefootnote{\dag}\footnotetext{Equal Contribution. Work done during an internship at Baidu.}
\def\thefootnote{*}\footnotetext{Corresponding author is Guanbin Li.}

\input{sec/0_abstract}    
\input{sec/1_intro}
\input{sec/2_method}
\input{sec/3_experiment}
\subsection*{Acknowledgments}
This work is supported in part by the National Key
R\&D Program of China (NO.~2024YFB3908503 and 2024YFB3908500), and in part by the
National Natural Science Foundation of China (NO.~62322608).
\input{sec/X_suppl}
{
     \small
     \bibliographystyle{ieeenat_fullname}
     \bibliography{main}
}


\end{document}

%% file: sec/0_abstract.tex
\begin{abstract}
Offline vectorized maps constitute critical infrastructure for high-precision autonomous driving and mapping services. Existing approaches rely predominantly on single ego-vehicle trajectories, which fundamentally suffer from viewpoint insufficiency: while memory-based methods extend observation time by aggregating ego-trajectory frames, they lack the spatial diversity needed to reveal occluded regions. Incorporating views from surrounding vehicles offers complementary perspectives, yet naive fusion introduces three key challenges: computational cost from large candidate pools, redundancy from near-collinear viewpoints, and noise from pose errors and occlusion artifacts.

We present OptiMVMap, which reformulates multi-vehicle mapping as a select-then-fuse problem to address these challenges systematically. An Optimal Vehicle Selection (OVS) module strategically identifies a compact subset of helpers that maximally reduce ego-centric uncertainty in occluded regions, addressing computation and redundancy challenges. Cross-Vehicle Attention (CVA) and Semantic-aware Noise Filter (SNF) then perform pose-tolerant alignment and artifact suppression before BEV-level fusion, addressing the noise challenge. This targeted pipeline yields more complete and topologically faithful maps with substantially fewer views than indiscriminate aggregation.
On nuScenes and Argoverse2, OptiMVMap improves MapTRv2 by +10.5 mAP and +9.3 mAP, respectively, and surpasses memory-augmented baselines MVMap and HRMapNet by +6.2 mAP and +3.8 mAP on nuScenes. These results demonstrate that uncertainty-guided selection of helper vehicles is essential for efficient and accurate multi-vehicle vectorized mapping. The code is released at https://github.com/DanZeDong/OptiMVMap.

\end{abstract}

%% file: sec/1_intro.tex
\section{Introduction}
\label{sec:Intro}
\begin{figure}[h]
\includegraphics[width=\columnwidth]{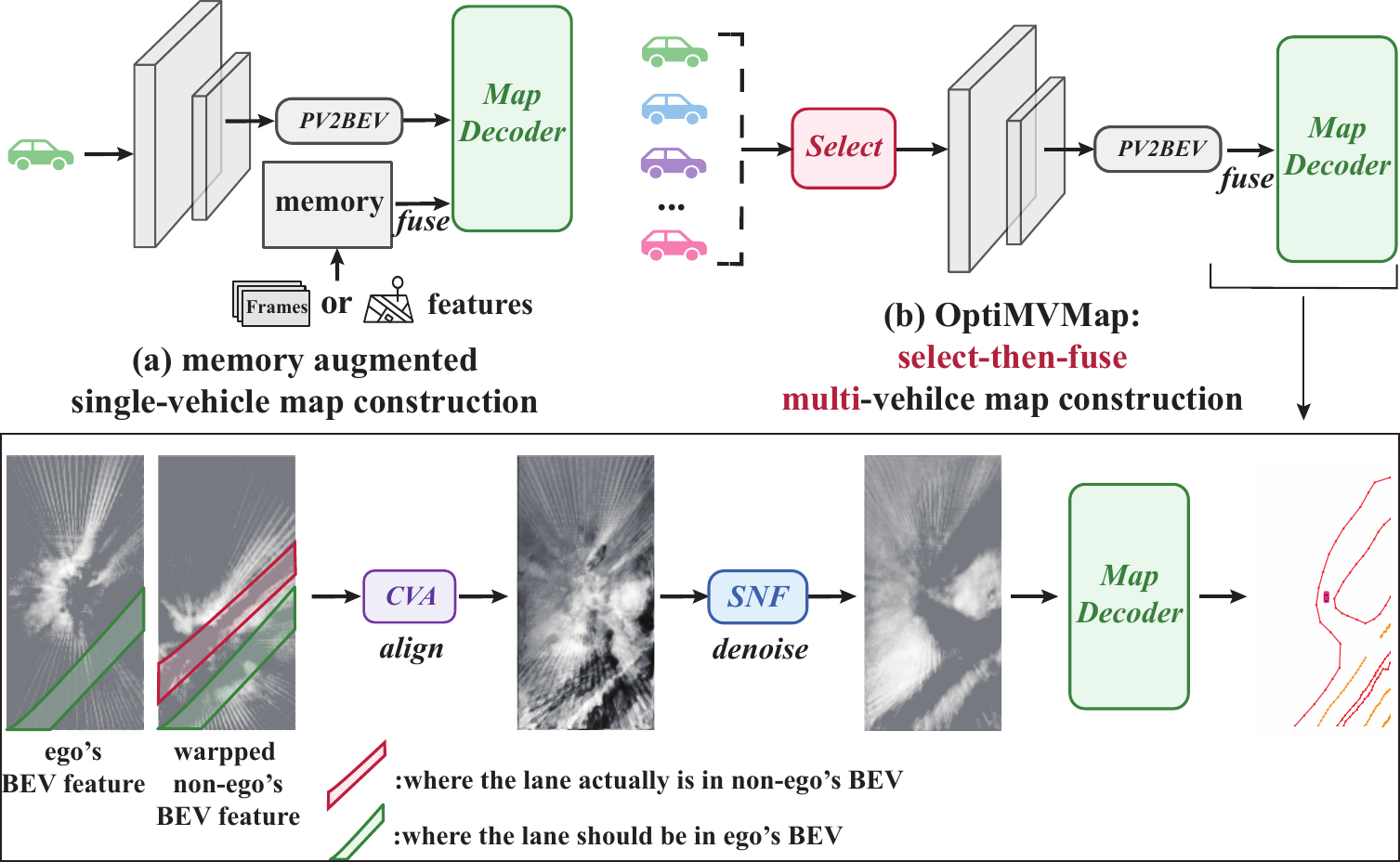}
\centering
\caption{
Comparison on the various design of map construction framework. (a) Memory-augmented single-vehicle methods with temporal memory (e.g.MVMap) or coarse map history (e.g.HRMapNet). (b) OptiMVMap: adopts Select–then–Fuse paradigm: selects complementary non-ego views first, then aligns and denoises during fusion.
}
\label{fig:motivation}
\end{figure}

Vectorized map construction—recovering structured map elements (lanes, boundaries, crossings) as geometric primitives from sensor observations—is critical infrastructure for autonomous driving, enabling localization, path planning, and decision-making \cite{chen2023vma, xia2024dumapnet, zhu2023nemo}. 
Current approaches fall into two paradigms: \textbf{online} methods decode maps frame-by-frame in real-time under strict latency constraints \cite{liao2023maptr, liao2024maptrv2, ding2023pivotnet, zhang2024gemap, zhang2024mapvr}, while \textbf{offline} methods collect sensor logs in-vehicle and generate maps off-board, eliminating communication and bandwidth constraints.
We focus on the offline setting—the de facto industry practice \cite{chen2023vma, xia2024dumapnet}—which adopts a ``collect first, build later'' paradigm for production mapping systems, and is broadly consistent with large-scale off-board auto-labeling pipelines in autonomous driving\cite{yang2025fusion4dal}. Existing offline methods predominantly rely on single-vehicle trajectories, formulating map construction as object detection and adopting DETR-based architectures to decode vectorized elements from ego-centric BEV features \cite{liu2023vectormapnet, liao2024maptrv2, ding2023pivotnet, Mask2Map} (Fig.~\ref{fig:motivation}(a)).
However, single-trajectory construction faces a fundamental bottleneck: \textbf{viewpoint insufficiency}—restricted to one perspective, the ego vehicle suffers from occlusions and long-range degradation, resulting in missing map elements in complex or distant areas.

To address the viewpoint insufficiency, recent methods adopt memory-augmented strategies in two primary forms (illustrated in Fig.~\ref{fig:motivation}b): (i) temporal memory aggregates adjacent frames from the same trajectory \cite{yuan2024streammapnet, zhu2024icmapper, liao2024maptrv2, xie2023mvmap}; and (ii) map-level history memory accumulates nearby outputs into a coarse, rasterized local map \cite{zhang2024hrmapnet}. While these methods expand temporal context, they remain fundamentally single-vehicle: the additional observations come from redundant and nearly collinear viewpoints, thus failing to introduce the diversity and complementarity necessary to reveal occluded or distant regions. 
Worse, without quality control, early errors propagate and corrupt future predictions. \emph{Critically, memory-based methods merely extend observation time but do not fundamentally resolve viewpoint insufficiency—they lack the spatial diversity that only complementary vehicle perspectives can provide}

\begin{figure}[t]
\includegraphics[width=\columnwidth]
{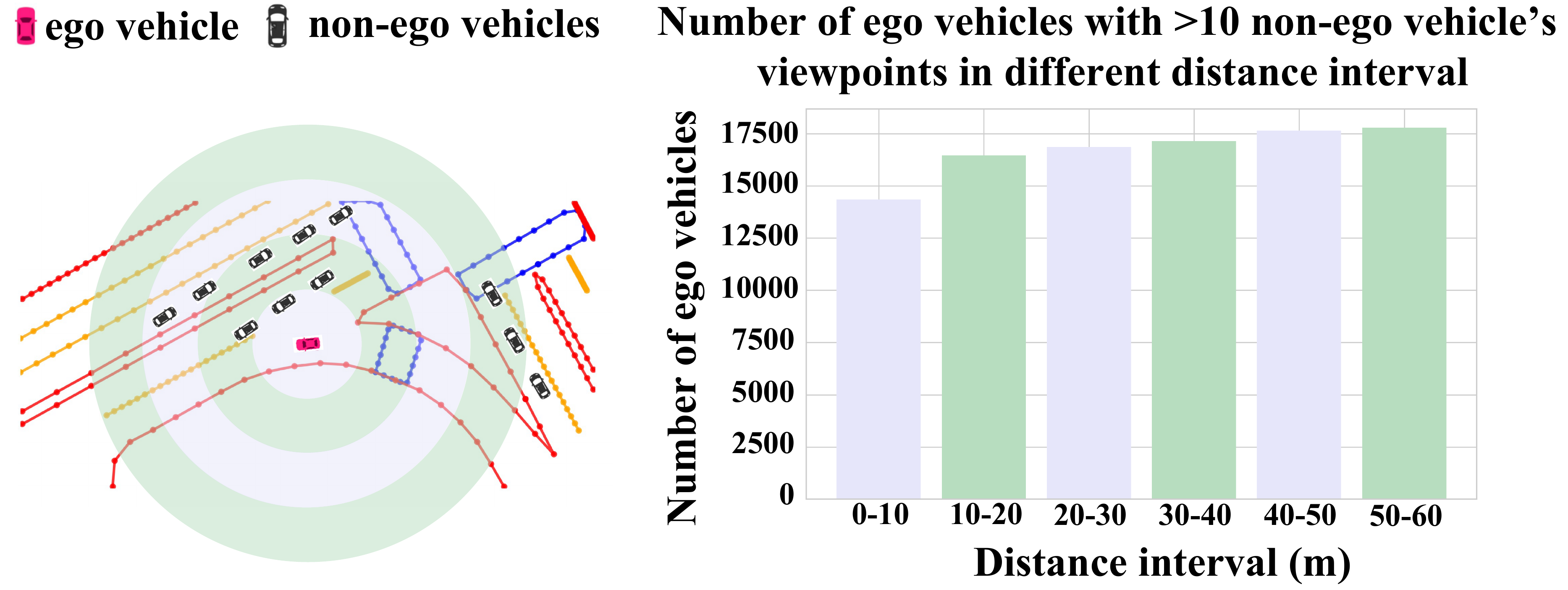}
\centering
\caption{Surrounding vehicle analysis. Across all distance intervals, there are more than 14,000 ego-vehicles with at least 10 non-ego vehicles in nuScenes dataset.
}
\label{fig:dist}
\end{figure}

An intuitive solution is to move beyond a single ego viewpoint and incorporate complementary views from other vehicles: larger inter-view parallax and time gaps restore visibility, reducing uncertainty in occluded and long-range regions. Indeed, our analysis reveals that 84.6\% (nuScenes) and 70.2\% (AV2) of scenes include other trajectories within 60m, and the ego vehicle is typically accompanied by 10+ non-ego viewpoints across distance bins (Fig.~\ref{fig:dist}), rendering multi-vehicle integration both feasible and promising. 
However, effectively leveraging this abundance of viewpoints is non-trivial. Naively fusing all nearby vehicles introduces three key challenges: (i) the candidate pool is large and heterogeneous, making exhaustive fusion computationally prohibitive; (ii) spatial proximity does not guarantee complementary information—nearby vehicles may provide redundant near-collinear views; and (iii) indiscriminate fusion amplifies noise from pose errors and occlusion-induced artifacts. To address these challenges, we adopt a \emph{select-then-fuse} paradigm: strategically identify a compact subset of helper vehicles that maximally reduce ego-centric uncertainty in occluded regions, then refine their alignment and suppress fusion noise before integration.

To instantiate this paradigm, we propose \textbf{Opt}imal \textbf{M}ulti-\textbf{V}ehicle Vectorized \textbf{Map} Construction (\textbf{OptiMVMap}), a plug-and-play framework (overview in Fig.~\ref{fig:model}). At its core lies an \textbf{Optimal Vehicle Selection (OVS)} module that transforms selection into a principled decision problem: it ranks non-ego candidates by their expected reduction in ego-centric BEV uncertainty over occluded and long-range regions while accounting for spatial relevance, geometric visibility, and viewpoint complementarity, retaining only a compact subset to bound computation and eliminate redundancy. Critically, even after selection, pose and temporal drift cause BEV misalignment (Fig.~\ref{fig:motivation}, bottom), and residual fusion noise persists. We address these through a lightweight refinement approach: \textbf{Cross-Vehicle Attention (CVA)} performs pose-tolerant BEV alignment to compensate for extrinsic inaccuracies and temporal asynchrony, while a \textbf{Semantic-aware Noise Filter (SNF)} applies learned, content-aware weights to suppress occlusion- and dynamics-induced artifacts. The refined BEV features are then fused, yielding maps that are more complete and topologically faithful than those produced by view-agnostic aggregation, yet obtained with substantially fewer fused views.

To enable multi-vehicle research, we extend nuScenes and Argoverse2 to nuScenes-MV and AV2-MV by associating each ego frame with helper views from other trajectories (60m radius, $\geq$30min separation). Evaluation protocols remain unchanged for fair comparison. Comprehensive evaluations demonstrate that OptiMVMap outperforms existing methods by substantial margins. When integrated with MapTRv2, it establishes a new state of the art, improving mAP over the single-vehicle baseline by \textbf{+10.5mAP} on nuScenes and \textbf{+9.3mAP} on Argoverse2. Furthermore, it surpasses the memory-augmented single-vehicle methods MVMap and HRMapNet by \textbf{+6.2 mAP} and \textbf{+3.8 mAP}, respectively, on nuScenes.

Our main contributions are as follows:
\begin{itemize}
\item To our knowledge, we are the first to systematically investigate offline vectorized map construction from a multi-vehicle perspective, identifying that indiscriminate view aggregation is suboptimal due to redundant viewpoints and pose/temporal noise.
\item We propose \textbf{OptiMVMap}, 
which reformulates multi-vehicle mapping as a \emph{select–then–fuse} problem to aggregate diverse yet complementary views. 
\item OptiMVMap integrates seamlessly with existing vectorized map perception methods and achieves state-of-the-art performance on nuScenes and Argoverse2 with a small OVS-selected helper set, confirming that selective, complementary-view fusion is essential for robust vectorized mapping.
\end{itemize}

\section{Related Works}
\subsection{Single-vehicle Vectorized Map Construction}
\textit{Single-vehicle Methods.} Classical maps rely on SLAM with costly human refinement \cite{shan2018lego, shan2020lio}. Recent works automate vectorized map construction from onboard sensors. HDMapNet \cite{li2022hdmapnet} produces BEV semantic segmentations with heuristic vectorization. Recent work also explores improving lane/centerline learning through generative BEV priors, e.g., LaneDiffusion \cite{wang2025lanediffusion}. VectorMapNet \cite{liu2023vectormapnet} is the first end-to-end framework that autoregressively decodes point sequences. Building on this line, DETR-style designs decode instances directly from BEV features \cite{liao2023maptr, ding2023pivotnet, liao2024maptrv2, zhang2024gemap, Mask2Map}.  
\textit{Memory-augmented Single-vehicle Methods.} To mitigate the limited field of view of a single vehicle, StreamMapNet and IC-Mapper aggregate temporally adjacent frames along the same ego trajectory \cite{yuan2024streammapnet, zhu2024icmapper}. HRMapNet \cite{zhang2024hrmapnet} writes nearby predictions into a coarse local raster canvas, while MVMap \cite{xie2023mvmap} fuses entire ego trajectory frames with a NeRF-style module. However, these memories remain fundamentally single-vehicle: 
additional observations are drawn from nearly collinear viewpoints and thus provide limited parallax to recover occluded or long-range structure; further, the lack of memory quality control allows early errors to be preserved and propagated. 
In contrast, \textbf{OptiMVMap} differs fundamentally: rather than accumulating all ego frames, it follows a \emph{select–then–fuse} paradigm, selecting \emph{non-ego} vehicles from nearby trajectories to provide complementary context where the ego view is uncertain, while filtering noisy or redundant observations.

\subsection{Online Vectorized Map Construction}
Most single-vehicle end-to-end methods run in an \textit{online} manner—decoding vectorized elements per frame from the current sensor stream \cite{liao2023maptr, ding2023pivotnet, liao2024maptrv2, zhang2024gemap, Mask2Map}. Streaming-memory variants enlarge the temporal window yet remain fundamentally single-vehicle \cite{yuan2024streammapnet, zhu2024icmapper}. 
In contrast, we target the \textit{offline} setting: leveraging cross-trajectory evidence from multiple vehicles to harvest complementary views and improve reliability under occlusions and at long range, while remaining compatible as a plug-in to standard online backbones.

\subsection{Offline Vectorized Map Construction}
Offline vectorized maps are the backbone of high-precision driving and mapping services \cite{chen2023vma, xia2024dumapnet}. VMA \cite{chen2023vma} proposes a scalable divide-and-conquer system that represents elements as unified point sequences for city-scale production. DuMapNet \cite{xia2024dumapnet} injects priors as queries with group-wise prediction and contextual prompts, improving spatial consistency and enabling large-scale deployment. LDMapNet-U \cite{xia2025ldmapnetu} further extends this production-oriented line to lane-level map updating with prior-map encoding and instance-level change prediction. 
Beyond these production systems, MVMap \cite{xie2023mvmap} performs dense volumetric fusion of an ego trajectory in an offline pipeline. 
Although this yields stronger spatial consistency than streaming approaches, it aggregates only \emph{single-vehicle} views and lacks utility-aware admission, leading to near-collinear redundancy and propagation of early errors. In contrast, our OptiMVMap adopts a select-then-fuse strategy guided by semantic/uncertainty cues, retaining only a small set of high-value candidates and focusing computation on complementary evidence.

%% file: sec/2_method.tex
\section{Methodology}

\begin{figure*}[!t]
    \centering
    \includegraphics[width=\linewidth]{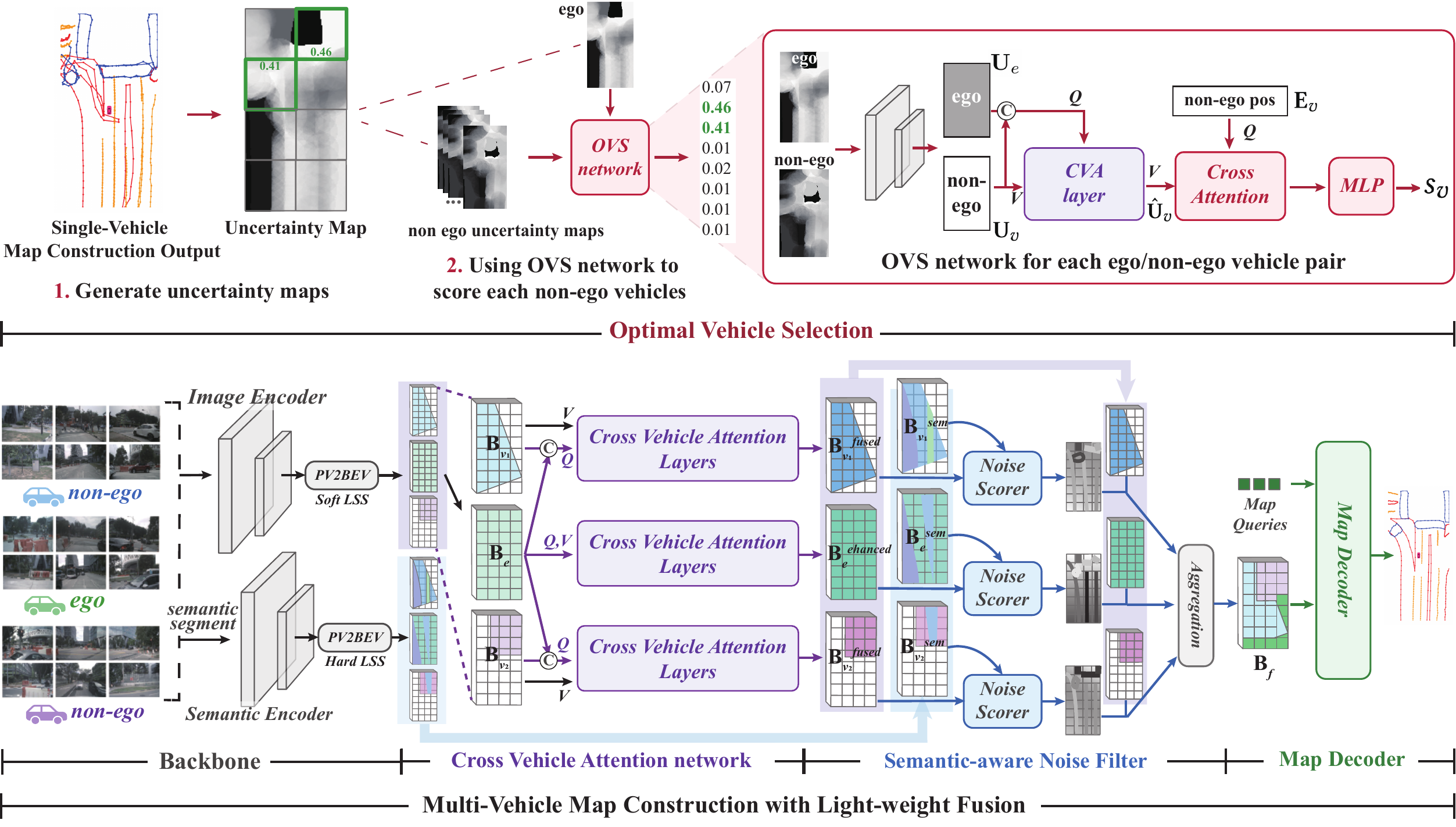}
    \caption{\textbf{Overview of OptiMVMap (Select then Fuse).} Firstly, OVS ranks non-ego vehicles by their expected reduction of ego-centric BEV uncertainty (occluded/long-range) and selects a compact top-K. Selected views are aligned via pose-tolerant Cross-Vehicle Attention (CVA), then denoised and aggregated by a Semantic-aware Noise Filter (SNF) into a fused BEV feature. A DETR-style decoder queries the fused BEV to produce vectorized map instances; the module is plug-and-play and detector-agnostic.}
    \label{fig:model}
\end{figure*}

\subsection{Problem Setup}
Given surround-view images of the ego vehicle $\mathcal{I}_{e}=\{\mathbf{I}^{i}_{e}\}_{i=1}^{N}$ and surround-view images of $M$ candidate non-ego vehicles $\{\mathcal{I}_{v_j}\}_{j=1}^{M}$ with $\mathcal{I}_{v_j}=\{\mathbf{I}^{i}_{v_j}\}_{i=1}^{N}$, our \emph{offline} setting considers cross-trajectory observations that may be asynchronous but can be temporally and spatially aligned to a common BEV frame. From the $M$ candidates, the OVS module selects a top-$K$ subset to serve as helper vehicles ($K\!\ll\!M$), where $N$ is the number of surround-view cameras per vehicle. The goal of offline multi-vehicle vectorized map construction is to recover the local BEV map from the ego images together with the selected non-ego evidence. The final output is a set of vectorized map elements $\{(c,\mathbf{P})\}$, where $c$ is a semantic class label and $\mathbf{P}=\{(x_i,y_i)\}_{i=1}^{L}$ is an ordered 2D point sequence of length $L$ in the BEV coordinate system.

\subsection{Overview}
\label{sec:overview}
In general, we adopt a \emph{select–then–fuse} offline pipeline with two stages: 
\textbf{(1) Optimal Vehicle Selection.} Given the ego and $M$ candidate non-ego vehicles, OVS ranks candidates by their \emph{expected reduction in ego-centric BEV uncertainty} over occluded and long-range regions—using spatial relevance, visibility, and baseline complementarity—and selects a compact top-$K$ set. 
\textbf{(2) Lightweight fusion.} The selected views enter a two-step BEV feature path (Fig.~\ref{fig:model}): \textbf{Cross-Vehicle Attention (CVA)} performs pose-tolerant cross-trajectory alignment and information exchange to produce per-pair aligned features; a \textbf{Semantic-aware Noise Filter (SNF)} then applies learned semantic weights to suppress occlusion-/dynamics-induced artifacts and aggregates the aligned features into a consolidated BEV representation. 
Finally, a standard DETR-style map decoder operates on the fused BEV feature to produce vectorized map instances. \emph{OptiMVMap is plug-and-play and detector-agnostic}: it attaches at the BEV feature level as a pre-decoder module and requires no changes to downstream decoders.

\subsection{Optimal Vehicle Selection}
\label{sec:selection}

Selecting non-ego vehicles optimally is central to reliable mapping: useful helpers are those that (i) most reduce the ego’s BEV uncertainty in occluded and long-range regions and (ii) occupy geometrically complementary positions relative to the ego. Guided by these factors, we introduce the \textbf{Optimal Vehicle Selection (OVS)} module to choose the most beneficial non-ego vehicles.

\textit{Two-stage strategy.} As illustrated in Fig.~\ref{fig:model}, OVS proceeds in two stages: \emph{single-vehicle uncertainty estimation} followed by \emph{multi-vehicle selection}. First, for each vehicle, we compute a BEV uncertainty map that highlights high-uncertainty areas; we then form the candidate set for each ego by spatial partitioning and proximity. Next, we create the candidate non-ego vehicle set for each ego vehicle according to their positions. The uncertainty maps and positions of both ego and candidate non-ego vehicles are then fed into the OVS network, which selects the optimal non-ego vehicles. By design, the budget $K$ is small (e.g., 1–3), preventing redundant, near-collinear choices and bounding downstream computation. This approach significantly reduces uncertainty within the ego vehicle’s critical perception regions. Finally, we integrate selected non-ego vehicles' views to obtain the final multi-vehicle vectorized map.

Firstly, to generate the uncertainty map for each vehicle, we individually input each vehicle’s views into the map construction model and obtain point-wise classification uncertainties $U_\mathbf{P}$ corresponding to each polyline point $\mathbf{P}$. 
These uncertainties indicate regions that need improvement \emph{without} non-ego assistance.
To obtain pixel-wise uncertainty $U_{i, j}$, 
we convert them to a pixel-wise BEV map by averaging within a radius-$d$ neighborhood:
\begin{equation}
\label{eq:unc}
    U_{i, j} = \frac{1}{|\boldsymbol{\Omega}_{i, j}|} \sum_{\mathbf{P} \in \boldsymbol{\Omega}_{i, j}} U_\mathbf{P}.
\end{equation}
where $\boldsymbol{\Omega}_{i,j}$ collects points within distance $d$ of pixel $(i,j)$.

Then, to create the candidate non-ego set, we partition the ego vehicle's BEV space into $N_{h}\times N_{w}$ equally-sized square regions, selecting the closest non-ego vehicle to each region center as a candidate.

Next, we encode pixel-wise uncertainty maps into compact features $\mathbf{U}_{e}$ (ego) and $\mathbf{U}_{v}$ (each non-ego $v$) via a small convolutional network $\mathcal{G}_{\theta}$:
\begin{equation}
\mathbf{U}_{e} = \mathcal{G}_{\theta}(U_{e}), \quad \mathbf{U}_{v} = \mathcal{G}_{\theta}(U_{v}).
\end{equation}
To assess uncertainty-reducing potential, we spatially align $\mathbf{U}_{e}$ with each candidate $\mathbf{U}_{v}$ using a CVA step, generating integrated uncertainty features:
\begin{equation}
\hat{\mathbf{U}}_v = \mathcal{F}(\mathbf{U}_{e}, \mathbf{U}_{v}).
\end{equation}
Each non-ego’s spatial position embedding $\mathbf{E}_{v}$ then serves as a query in a cross-attention module with $\hat{\mathbf{U}}_v$, and an MLP produces a suitability score $s_{v}$:
\begin{equation}
s_{v} = \text{MLP}(\textrm{CA}(\mathbf{E}_{v}, \hat{\mathbf{U}}_v)).
\end{equation}
Finally, we select the top-$K$ non-ego vehicles based on $\{s_{v}\}$. By construction, OVS concentrates computation on high-utility, geometrically complementary helpers and targets uncertainty reduction in the ego’s hard regions, providing the inputs for the subsequent CVA/SNF fusion pipeline.

\subsection{Multi-vehicle Vectorized Map Construction}
\textbf{BEV Encoder.}
We first extract per-view (PV) \emph{image} and \emph{semantic} features for the ego and each non-ego vehicle. Let the ego’s multi-view images and semantic segmentations be $\mathcal{I}_e$ and $\mathcal{S}_e$ (and analogously $\mathcal{I}_v,\mathcal{S}_v$ for a non-ego $v$).
Using separate convolutional image and semantic encoders, we first derive PV features $\mathbf{F}_e$ and $\mathbf{F}_v$, and their semantic counterparts $\mathbf{F}_e^{sem}$ and $\mathbf{F}_v^{sem}$. These PV features are then transformed into BEV features $\mathbf{B}_e$ and $\mathbf{B}_v$, and their semantic counterparts $\mathbf{B}_e^{sem}$ and $\mathbf{B}_v^{sem}$, using the PV2BEV module.

\begin{figure}
    \centering
    \includegraphics[width=\columnwidth]{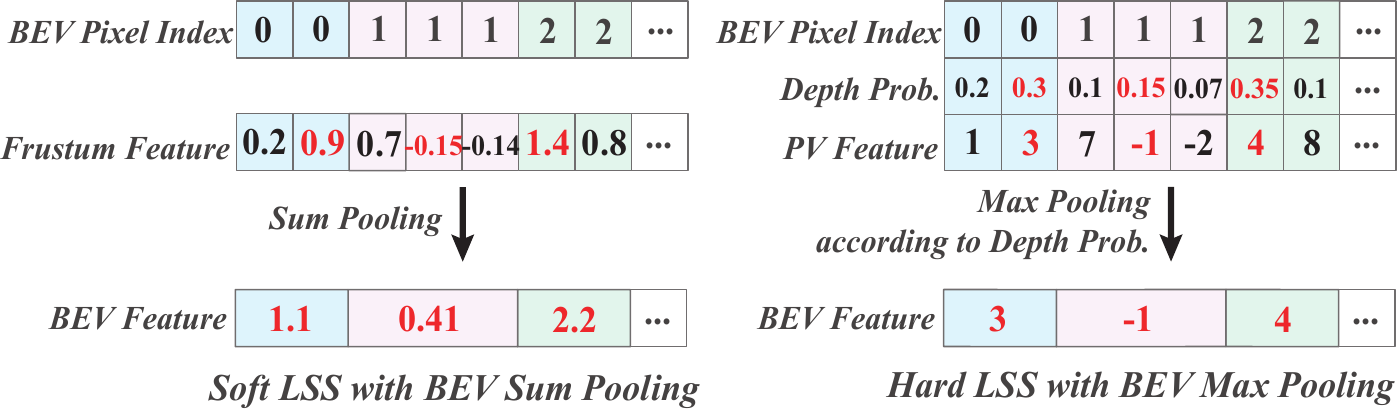}
    \caption{Comparison between Soft LSS and Hard LSS.}
    \label{fig:bevpooling}
\end{figure}
We use LSS \cite{philion2020lift} to transform PV image features into BEV features. In LSS's BEV pooling, it directly sums depth-weighted PV features (frustum features) into BEV grids, performing a soft assignment (Soft-LSS). However, for PV \textit{semantic} features, this soft assignment leads to semantic ambiguity. To mitigate this, we propose Hard-LSS: we directly assign PV semantic features to BEV grids and use depth probabilities as confidence scores. As illustrated in Fig.~\ref{fig:bevpooling}, for each BEV grid cell, we retain only the semantic feature corresponding to the highest depth probability. Each non-ego BEV feature $\mathbf{B}_v$ is then initially aligned to the ego BEV coordinates by BEV warping using coordinate transformations derived from camera parameters.

\noindent\textbf{Cross-Vehicle Attention (CVA) network.}
\label{sec:featurefusion}
We apply CVA only to the OVS-selected helpers so that cross-vehicle alignment and fusion remain lightweight. 
The Cross-Vehicle Attention (CVA) network progressively aligns and integrates BEV features from ego and non-ego vehicles. Due to calibration errors across vehicles, BEV warping cannot fully resolve rotational and translational misalignments. 
CVA addresses these by \emph{learned, feature-conditioned sampling} while exchanging information across vehicles.
A CVA layer is denoted as $\mathcal{F}(\mathbf{Q}_{in}, \mathbf{V})$. First, each non-ego BEV feature $\mathbf{B}_v$ is individually fused with the ego feature $\mathbf{B}_e$, generating intermediate aligned features. Subsequently, another CVA layer further refines alignment by integrating these intermediate features with the original non-ego features:
\begin{equation}
\mathbf{B}_v^{fused} = \mathcal{F}(\mathcal{F}(\mathbf{B}_e, \mathbf{B}_v), \mathbf{B}_v),
\end{equation}
where $\mathbf{B}_v^{fused}$ denotes the refined non-ego BEV feature after alignment and integration.
Specifically, a CVA layer $\mathcal{F}$ is adapted from deformable attention \cite{zhu2021deformable}. Given a query feature $\mathbf{Q}_{in}$, value feature $\mathbf{V}$, and a set of reference points $\mathbf{R}$, CVA computes adaptive sampling offsets and weights as follows:
\begin{equation}
\mathbf{Q}_{out} = \mathbf{Q}_{in} + \sum_{i=1}^{N_{off}} \mathbf{W}_i \cdot \textrm{DA}(\mathbf{Q}_{in}, \mathbf{R}+\mathbf{O}_i, \mathbf{W}_v\mathbf{V}),
\end{equation}
\begin{equation}
\mathbf{O}_i = \textrm{OffsetProj}([\mathbf{Q}_{in}, \mathbf{V}]), \quad
\mathbf{W}_i = \textrm{WeightEmbed}([\mathbf{Q}_{in}, \mathbf{V}]),
\end{equation}
where $\mathbf{O}_i$ and $\mathbf{W}_i$ represent sampling offsets and weights, respectively; $N_{off}$ is the number of sampling offsets for each query; and $\mathbf{R}$ denotes reference points. We use two shallow CVA layers with a small $N_{off}$, yielding near-linear cost in $HW$ and $K$ under our small-$K$ regime.
Additionally, to ensure consistency, the ego vehicle's BEV features are also self-enhanced using CVA:
\begin{equation}
\mathbf{B}_e^{enhanced} = \mathcal{F}(\mathcal{F}(\mathbf{B}_e, \mathbf{B}_e), \mathbf{B}_e).
\end{equation}

\noindent\textbf{Semantic-aware Noise Filter (SNF).}
After CVA, residual artifacts may persist (e.g., dynamics- and occlusion-induced noise).
Guided by semantic and uncertainty priors, SNF computes pixel-wise weights indicating confidence in each feature’s reliability. SNF acts as a quality gate, normalizing contributions across $\text{ego} + \text{selected non-ego}$ vehicles at each pixel. Weights for $\mathbf{B}_e^{enhanced}$ and each $\mathbf{B}_{v_j}^{fused}$ are computed as:
\begin{equation}
\begin{aligned}
\mathbf{S}_{e}, \mathbf{S}_{v_1},...,\mathbf{S}_{v_K} =
  \textrm{Softmax}(& \textrm{NS}(\mathbf{B}_e^{enhanced}, \mathbf{B}_e^{sem}), \\
& \textrm{NS}(\mathbf{B}_{v_1}^{fused}, \mathbf{B}_{v_1}^{sem}),..., \\
& \textrm{NS}(\mathbf{B}_{v_K}^{fused}, \mathbf{B}_{v_K}^{sem}))
\end{aligned}
\end{equation}
where NS is a noise-scoring convolutional network. 
The final BEV representation is an adapted combination of the ego-enhanced and fused non-ego features, modulated by the learned reliability weights:
\begin{equation}
\mathbf{B}_f
= \mathbf{S}_{e} \odot \mathbf{B}^{\mathrm{enhanced}}_{e}
+ \sum_{j=1}^{K} \mathbf{S}_{v_j} \odot \mathbf{B}^{\mathrm{fused}}_{v_j},
\end{equation}
where $\odot$ denotes element-wise multiplication. This semantic-aware gating suppresses noisy or conflicting evidence while stabilizing fusion across sources.

\noindent\textbf{Map decoder.}
\label{sec:decoder}
Given the fused BEV feature $\mathbf{B}_{f}$ produced by CVA and SNF, we employ a standard vectorized map decoder to generate the vectorized map. The framework is decoder-agnostic: any DETR-style design and query-initialization scheme can be used (e.g., MapTRv2 \cite{liao2024maptrv2}) without modification.

\subsection{Training}
\label{sec:training}
We train in two stages. (i) \emph{Fusion backbone pretraining:} CVA\,+\,SNF\,+\,decoder are trained with random non-ego vehicle sampling (OVS disabled). (ii) \emph{OVS training:} the pretrained fusion backbone is fixed and OVS is learned to select helpers. At inference, only the OVS-selected Top-$K$ views are fused, preserving the lightweight design.
To train the multi-vehicle vectorized map construction model, following MapTRv2 \cite{liao2024maptrv2}, we use a one-to-one set prediction loss including a classification loss, a point-to-point loss and an edge direction loss. When plugging OptiMVMap into MapTRv2-like methods, the one-to-many set prediction loss $\mathcal{L}_{one2many}$ is also adopted. Besides, we use a dense prediction loss $\mathcal{L}_{dense}$ to further leverage semantic and geometric information, consisting of a depth prediction loss, a BEV semantic segmentation loss, a BEV instance segmentation loss and a PV segmentation loss. Note that $\mathcal{L}_{dense}$ is applied on $\mathbf{B}_e^{enhanced}$, $\{\mathbf{B}_{v_j}\}_{j=1}^{K}$ and $\mathbf{B}_{f}$. The overall loss is:
\begin{equation}
\mathcal{L}_{map} = \beta_{o}\mathcal{L}_{one2one} + \beta_{m}\mathcal{L}_{one2many} + \beta_{d}\mathcal{L}_{dense}.
\end{equation}

To stabilize OVS and avoid label ambiguity, we construct candidate sets such that the non-ego vehicle selected from each of the $8$ BEV regions is distinct, yielding a unique optimal subset for the chosen $K$.
Every possible vehicle combination is evaluated via mean average precision (mAP), and the highest-score set is used as ground truth. We train OVS with a sigmoid BCE loss:
\begin{equation}
\mathcal{L}_{\text{OVS}} = -\frac{1}{|\mathcal{V}|} \sum_{v \in \mathcal{V}} \left[ y_v \log(\sigma(s_v)) + (1 - y_v) \log(1 - \sigma(s_v)) \right],
\end{equation}
where $\mathcal{V}$ is the set of candidate non-ego vehicles and $\sigma(\cdot)$ denotes the sigmoid function. This objective drives the network to assign higher scores to candidate views that effectively reduce uncertainty and improve mapping quality.

%% file: sec/3_experiment.tex
\section{Experiments}
\subsection{Experimental Settings}
\textbf{Datasets.} The large-scale nuScenes\cite{caesar2020nuscenes} dataset includes 2D city-level global vectorized maps and 1,000 scenes, each approximately 20 seconds long, with each frame providing RGB images from 6 cameras. The Argoverse 2 (AV2)\cite{wilson2023argoverse2} dataset contains 1000 scenes from 6 U.S. cities, each roughly 15s long, and provides synchronized RGB images from seven cameras. As introduced in Sec.\ref{sec:Intro}, we extend the two datasets into \textbf{nuScenes-MV} and \textbf{Argoverse2-MV} datasets as follows: for each ego vehicle, we first select non-ego vehicles within a 60m radius from other trajectories with a $\geq30$min time separation to avoid near-collinear viewpoints and to leverage occlusion changes; If none exist, we select frames from the same trajectory as fallback.

\noindent\textbf{Metrics.} We conduct evaluation with Chamfer distance based Average Precision (AP) following previous mainstream works \cite{li2022hdmapnet, liao2024maptrv2} for fair comparisons. The AP is calculated under the average of three Chamfer distance thresholds of 0.5, 1.0, and 1.5 meters. The perception ranges are [-15.0m, 15.0m] for the X-axis and [-30.0m, 30.0m] for the Y-axis. Three types of map instances are selected for HD map construction, including pedestrian crossing, lane divider, and road boundary. 

\noindent\textbf{Implementation Details.} 
Please refer to the \underline{supplementary document} for more details.

\begin{table*}[t]
\begin{center}
\caption{Comparison with SOTA on nuScenes. Unless otherwise specified, we use $K{=}2$ non-ego helpers for OptiMVMap plug-in experiments.}
\label{tab:nuscenes}
\renewcommand{\arraystretch}{1.0}
\resizebox{0.75\textwidth}{!}{
\begin{tabular}{c|c|c|cccc} 
\midrule
Methods & Backbone & Epochs & $AP_{\textit{div}}(\uparrow)$ & $AP_{\textit{ped}}(\uparrow)$ & $AP_{\textit{bnd}}(\uparrow)$ & mAP$(\uparrow)$
\\
\midrule
MapTR \cite{liao2023maptr} & R50 &24 & 46.3 & 51.5 & 53.1 & 50.3 
\\
MapVR \cite{zhang2024mapvr} & R50 &24 & 56.2 & 56.5 & 60.1 & 57.6 
\\
PivotNet \cite{ding2023pivotnet} & R50 &30 & 47.7 & 54.4 & 51.4 & 51.2 
\\
BeMapNet \cite{qiao2023bemapnet} & R50 &30 & 62.3 & 57.7 & 59.4 & 59.8 
\\
StreamMapNet \cite{yuan2024streammapnet} & R50 &30 & 66.3 & 61.7 & 62.1 & 63.4 
\\
HIMap \cite{zhou2024himap} &R50 &30 &68.4 &62.6 &69.1 &66.7
\\
Mask2Map \cite{Mask2Map} &R50 &24 &71.3 &70.6 &72.9 &71.6
\\
InteractionMap \cite{wu2025interactionmap} &R50 &24 &\textbf{72.7} &69.7 &73.0 &71.8
\\
\midrule
VectorMapNet \cite{liu2023vectormapnet} & R50 &24 & 36.1 & 47.3 & 39.3 & 40.9
\\
VectorMapNet + MVMap \cite{xie2023mvmap} & R50 &24 & 55.0 & 46.2 & 45.5 & 48.9
\\
VectorMapNet + OptiMVMap & R50 &24 & 51.2 & 60.3 & 54.8 & 55.1 
\\
\midrule
MGMap \cite{liu2024mgmap} & R50 &24 & 61.8 & 65.0 & 67.5 & 64.8
\\
MGMap\dag \cite{liu2024mgmap} & R50 &24 & 64.2 & 60.1 & 65.3 & 63.2 
\\
MGMap + OptiMVMap & R50 &24 & 69.3 & 66.5 & 72.2 & 69.3
\\
\midrule
MapTRv2 \cite{liao2024maptrv2} & R50 &24 & 59.8 & 62.4 & 62.4 & 61.5 
\\
MapTRv2 + HRMapNet \cite{zhang2024hrmapnet} & R50 &24 & 67.4 & 65.8 & 68.5 & 67.2
\\
MapTRv2 + OptiMVMap & R50 &24 & 70.0 & 70.6 & 72.2 & 71.0
\\
MapTRv2 + OptiMVMap + QI & R50 &24 & 71.0 & 71.5 & 73.6 & 72.0  
\\
MapTRv2 + OptiMVMap ($K$=5) & R50 &24 & \underline{72.5} & \underline{73.8} & \underline{75.4} & \underline{73.9}
\\
MapTRv2 + OptiMVMap ($K$=7) & R50 &24 & \textbf{72.7} & \textbf{74.2} & \textbf{76.1} & \textbf{74.3}
\\
\midrule
MapTracker \cite{chen2024maptracker} & R50 &100 & 72.4 & 77.3 & 74.2 & 74.7 \\
MapExpert \cite{zhang2025mapexpert} & R50 &100 & \underline{73.9} & \textbf{79.4} & \underline{76.2} & \underline{76.5} \\
MapTRv2 + OptiMVMap ($K$=7) & R50 &110 & \textbf{76.1} & \underline{78.5} & \textbf{79.7} & \textbf{78.1}
\\
\midrule
\end{tabular}}
\end{center}
\vspace{-20pt}
\end{table*}

\begin{table}[t]
\begin{center}
\caption{Comparison with SOTA on AV2. Unless otherwise specified, we use $K{=}2$ non-ego helpers for OptiMVMap plug-in experiments. All results are obtained under the same setting with an ResNet50 backbone, 3-dim prediction.}
\label{tab:av2}
\renewcommand{\arraystretch}{1.0}
\resizebox{1.0\linewidth}{!}{
\begin{tabular}{c|cccc} 
\midrule
Methods 
& $AP_{\textit{div}}(\uparrow)$ & $AP_{\textit{ped}}(\uparrow)$ & $AP_{\textit{bnd}}(\uparrow)$ & mAP$(\uparrow)$ 
\\
\midrule
MapTRv2 \cite{liao2024maptrv2} & 69.3 & 59.3 & 64.4 & 64.3 
\\
HIMap \cite{zhou2024himap} &68.3 &66.7 &70.3 &68.4
\\
MapTracker \cite{chen2024maptracker} &66.4 &74.5 &73.4 &71.4
\\
InteractionMap \cite{wu2025interactionmap} &75.5 &67.7 &73.1 &72.1
\\
MapExpert \cite{zhang2025mapexpert} &66.9 &76.4 &75.1 &72.8
\\
MapTRv2 + HRMapNet \cite{zhang2024hrmapnet} & 71.5 & 64.1 & 69.7 & 68.5
\\
MapTRv2 + OptiMVMap & \textbf{75.9} & \textbf{70.4} & \textbf{74.7} & \textbf{73.6}
\\
\midrule
\end{tabular}}
\vspace{-20pt}
\end{center}
\end{table}
 
\subsection{Main Results}
\noindent\textbf{Select–then–Fuse delivers state-of-the-art with few helpers.}
Unless otherwise specified, we use a small helper $K{=}2$ non-ego vehicles.
On nuScenes (Table~\ref{tab:nuscenes}), simply plugging OptiMVMap into MapTRv2 already yields a substantial gain, and adding MGMap’s query initialization (QI) further lifts the improvement to +10.5 mAP (61.5$\rightarrow$72.0).
With few more budget ($K{=}5$), MapTRv2 + OptiMVMap reaches 73.9 mAP, setting a new \textbf{SOTA} on nuScenes (+2.1 over InterationMap on 24 epoch setting). For 110 epoch setting, we reach 78.1 mAP (+1.6 over MapExpert). We also surpass the prior best on AV2 by 0.8 mAP with only $K{=}2$.

\noindent\textbf{Plug-and-play across architectures.}
OptiMVMap acts as a modular layer that consistently improves diverse backbones.
For the autoregressive VectorMapNet, it brings +14.2 mAP (40.9$\rightarrow$55.1).
For the leading approach MGMap, OptiMVMap improves our reproduction MGMap by +6.1 mAP (63.2$\rightarrow$69.3), demonstrating that uncertainty-guided selection, pose-tolerant alignment, and semantic-aware denoising transfer effectively across decoding paradigms.

\noindent\textbf{Beyond memory-augmented single-vehicle.}
Compared with memory-augmented baselines, OptiMVMap achieves higher accuracy with only $K{=}2\!\sim\!3$ helpers.
Specifically, relative to MVMap (on VectorMapNet) we observe +6.2 mAP (48.9$\rightarrow$55.1), and relative to HRMapNet (on MapTRv2) we gain +3.8 mAP (67.2$\rightarrow$71.0).
These results support the premise in Sec.~\ref{sec:Intro}: complementary, non-collinear viewpoints selected by OVS, aligned by CVA, and denoised by SNF are more effective than longer temporal stacks or coarse map histories.
By fusing BEV features rather than output rasters, our pipeline preserves richer BEV information prior to decoding, yielding cleaner topology under occlusions and at long range.

\noindent\textbf{Cross-dataset generalization.}
On Argoverse 2 (Table~\ref{tab:av2}), plugging OptiMVMap into MapTRv2 improves the baseline by +9.3 mAP (64.3$\rightarrow$73.6) and surpasses HRMapNet by +5.1 mAP (68.5$\rightarrow$73.6), indicating that the uncertainty-guided select–then–fuse paradigm generalizes beyond nuScenes with the small helper budget. Achieving a new SOTA on AV2 (+1.5 over the InteractionMap, +0.8 over the MapExpert) further confirms OptiMVMap's cross-dataset generalization.

\begin{figure}[t]
    \centering
    \includegraphics[width=\linewidth]{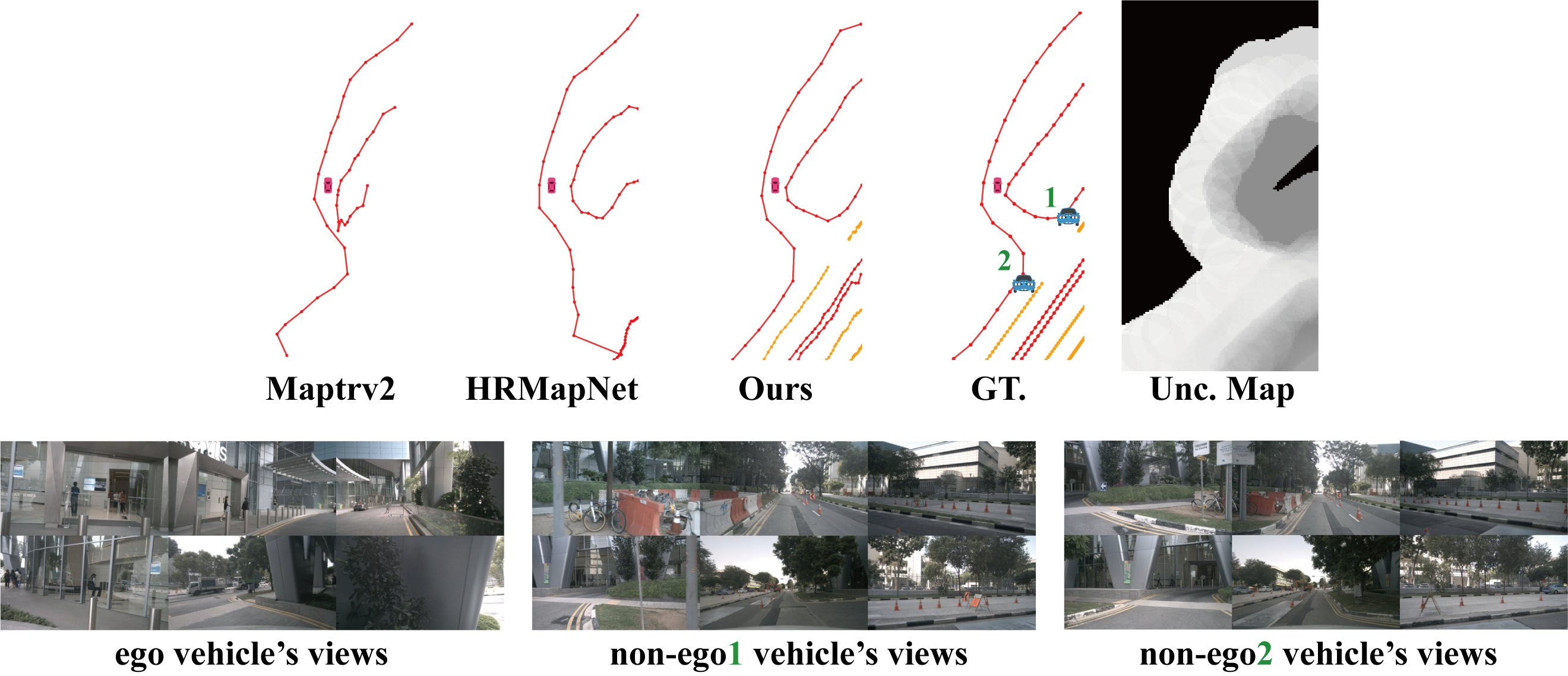}
    \caption{Comparison of qualitative results on the nuScenes dataset. ``Unc. Map" means uncertainty map. The blue car logos in GT. figure show the locations of non-ego vehicles.}
    \label{fig:vis-res}
    \vspace{-15pt}
\end{figure}

\subsection{Ablation Studies}
We conduct ablations to test the hypothesis that selecting a small, complementary helper set \emph{before} fusion is the primary driver of performance. Unless otherwise specified, all ablations are trained for 24 epochs on the full nuScenes training set. For the main component analysis, following Mask2Map, we train for 24 epochs on a \(1/4\) nuScenes subset to enable faster iteration.

\noindent\textbf{Contribution of Main Components.}
Table~\ref{tab:ablation} progressively adds each component to MapTRv2 baseline. Simply concatenating ego and non-ego features and fusing them through shallow MLPs provides only a marginal +1.8 mAP gain, since naive fusion fails to compensate for noise and cross-vehicle misalignment.
Adding CVA yields +2.6 mAP (39.5$\rightarrow$42.1), confirming that pose-tolerant cross-vehicle alignment is a prerequisite for reliable fusion.
Introducing SNF provides another +2.0 mAP (42.1$\rightarrow$44.1) by suppressing occlusion-/dynamics-induced artifacts.
Further replacing BEV \emph{sum} pooling with \emph{depth–semantic decoupled max pooling} brings +0.9 mAP (44.1$\rightarrow$45.0) for avoiding semantic blurring.
Finally, adding OVS on top lifts performance by a further +4.8 mAP (45.0$\rightarrow$49.8), exceeding the combined gains of alignment and denoising, indicating that selecting a small set of complementary helpers before fusion is the dominant driver of accuracy.

\noindent\textbf{Effect of non-ego vehicle selection.}
Table~\ref{tab:selection} compares \textbf{OVS} with two policies under the same helper budget $K$: 
(i) \emph{Random} — uniformly sampling $K$ helpers from the 60\,m candidate pool, averaged over 10 independent trials; 
(ii) \emph{Closest} — choosing the $K$ spatially nearest helpers in the ego-centric BEV frame.
The \emph{Random} average remains substantially lower, implying that indiscriminate aggregation of available frames tends to inject pose/asynchrony and occlusion noise and is therefore suboptimal.
Meanwhile, \emph{Closest} mirrors temporal stacking behaviors (e.g., selecting near-collinear views as in trajectory-wise fusion), which provides limited parallax and high redundancy, yielding smaller improvements.
\textbf{OVS} outperforms both by a clear margin (\textbf{+6.9} mAP over \emph{Random}, \textbf{+4.4} mAP over \emph{Closest}), indicating that selecting complementary viewpoints—rather than merely nearby ones—drives the gains.

\noindent\textbf{Analysis on helper budget $K$.}
As shown in Table~\ref{tab:nvnum}, increasing $K$ from 1$\rightarrow$2 yields \textbf{+4.8} mAP. Enlarging $K$ to 3, 4, and 5 brings only about \textbf{+1} mAP per step (71.0$\rightarrow$72.1$\rightarrow$72.9$\rightarrow$73.9), and beyond that the curve gradually flattens, with merely \textbf{+0.3} from $K{=}5$ to $6$ and \textbf{+0.1} from $K{=}6$ to $7$.
This small-$K$ knee shows that an OVS-selected set of \textbf{2-5} complementary helpers captures most of the attainable gains; simply accumulating more frames delivers diminishing returns and increases redundancy/pose–async noise—i.e., indiscriminately “fusing-all” is not optimal.

\noindent\textbf{Noise robustness under pose perturbations.}
Please refer to the \underline{supplementary document} for more details.

\begin{table}
\centering
\caption{Ablation analysis of key components on nuScenes.}
\vspace{-5pt}
\label{tab:ablation}
\setlength{\tabcolsep}{3pt}        
\renewcommand{\arraystretch}{1.0}  
\resizebox{0.9\linewidth}{!}{
\begin{tabular}{p{0.32\linewidth}cccc}
\toprule
Methods & AP$_{\textit{div}}$($\uparrow$) & AP$_{\textit{ped}}$($\uparrow$) & AP$_{\textit{bnd}}$($\uparrow$) & mAP($\uparrow$) \\
\midrule
MapTRv2 baseline &35.1 &34.8 &43.1 &37.7 \\
Naive Fusion & 38.0 & 35.0 & 45.5 & 39.5 \\
CVA & 40.5 & 38.7 & 47.1 & 42.1 \\
CVA + SNF & 43.9 & 38.2 & 50.2 & 44.1 \\
\makecell{CVA + SNF + \\BEV Max Pooling} & 42.8 & 41.2 & 50.9 & 45.0 \\
\makecell{CVA + SNF + \\Pooling + OVS} & 46.8 & 46.6 & 56.0 & 49.8 \\
\bottomrule
\end{tabular}}
\end{table}

\begin{table}
\begin{center}
\caption{Comparison of non-ego vehicle selection method (Random is averaged over 10 trials). }
\vspace{-5pt}
\label{tab:selection}
\renewcommand{\arraystretch}{1.0}
\resizebox{0.8\linewidth}{!}{
\begin{tabular}{c|ccccc} 
\midrule
Methods
& $AP_{\textit{div}}(\uparrow)$
& $AP_{\textit{ped}}(\uparrow)$
& $AP_{\textit{bnd}}(\uparrow)$
& mAP$(\uparrow)$
\\ 
\midrule
Random & 65.8 & 63.3 & 66.1 & 65.1 
\\
Closest & 68.3 & 65.6 & 68.5 & 67.6 
\\
OVS & \textbf{71.0} & \textbf{71.5} & \textbf{73.6} & \textbf{72.0}
\\
\midrule
\end{tabular}
}
\end{center}
\vspace{-10pt}
\end{table}

\begin{table}
\begin{center}
\caption{Analysis on non-ego vehicle number $K$.}
\vspace{-5pt}
\label{tab:nvnum}
\renewcommand{\arraystretch}{1.0}
\resizebox{0.7\linewidth}{!}{
\begin{tabular}{c|cccc} 
\midrule
$K$
& $AP_{\textit{div}}(\uparrow)$
& $AP_{\textit{ped}}(\uparrow)$
& $AP_{\textit{bnd}}(\uparrow)$
& mAP$(\uparrow)$
\\ 
\midrule
1 & 66.2 & 64.4 & 68.1 & 66.2 
\\
2 & 70.0 & 70.6 & 72.2 & 71.0  
\\
3 & 71.0 & 71.6 & 73.6 & 72.1 
\\
4 & 71.7 & 72.6 & 74.5 & 72.9
\\
5 & 72.5 & 73.8 & 75.4 & 73.9
\\
6 & 72.7 & 74.1 & 75.9 & 74.2
\\
7 & 72.7 & 74.2 & 76.1 & 74.3
\\
\midrule
\end{tabular}}
\end{center}
\vspace{-10pt}
\end{table}

\subsection{Qualitative Analysis}
In Fig.~\ref{fig:vis-res}, walls occlude the pavement behind the ego and the far-range straight lanes, where the ego’s uncertainty map peaks. 
The MapTRv2 baseline misses these distant centerlines and warps boundaries; HRMapNet aggregates such errors without quality control. 
OptiMVMap selects two non-ego helpers whose viewpoints look through the wall-occluded corridor and down the long straight, and—after alignment/denoising—recovers the straight lane lines and cleans the fork boundaries. 
The uncertainty hotspots precisely coincide with these regions and guide OVS to the most complementary views, explaining why selective fusion succeeds where indiscriminate aggregation fails. More qualitative results are in the supplementary material.

\section{Conclusion}
We revisit offline vectorized map construction from a multi-vehicle perspective and present OptiMVMap, a plug-and-play select–then–fuse framework that aggregates complementary views from surrounding vehicles. With only 2–5 selected helpers, OptiMVMap yields more complete and topologically faithful maps, particularly under occlusion and at long range. Integrated with MapTRv2, it achieves
state-of-the-art performance on nuScenes and Argoverse2, consistently surpassing strong single-vehicle and memory-augmented baselines. These results demonstrate that principled, complementary-view selection—rather than indiscriminate aggregation—is essential for robust offline vectorized mapping. We anticipate that our work will inspire future research on multi-perspective, selection-aware mapping approaches.

%% file: sec/X_suppl.tex
\clearpage
\setcounter{page}{1}
\maketitlesupplementary

\section{Implementation Details}
We employ ResNet50 \cite{he2016resnet} as the backbone network for image processing and ResNet18 for uncertainty map feature extraction in OVS. We utilize InternImage \cite{wang2023internimage} as the pretrained model for semantic prior generation. 
The default settings for instance queries, point queries, and decoder layers are 50, 20, and 6, respectively. We use the AdamW optimizer with a learning rate of $6 \times 10^{-4}$ and a weight decay of 0.01. All models are trained using 8 80GB NVIDIA Tesla A100 GPUs, with a batch size of 4 per node, leading to a total batch size of 32.  
For the overall loss, we set $\beta_{o} = 1$, $\beta_{m} = 1$, $\beta_{d} = 1$, $\beta_{c} = 1$.

\section{Noise robustness under pose perturbations.}
Tab.~\ref{tab:noise} injects Gaussian rotation/translation noise into vehicle poses. From 0 to 0.05rad rotation noise, the MapTRv2 baseline drops -29.2 mAP, whereas CVA retains most accuracy (-5.3 mAP).
At 1m translation std., the baseline collapses to 18.8 mAP while our model still achieves 62.3 mAP.
These show that explicit cross-vehicle alignment via CVA is essential for making BEV fusion robust to extrinsic inaccuracies. 

\begin{table}[t]
\centering
\caption{Robustness of injected pose noise on nuScenes. 
We report mAP under varying standard deviations of Gaussian rotation/translation noise.
}
\label{tab:noise}
\resizebox{0.8\linewidth}{!}{
\begin{tabular}{p{0.28\linewidth}ccccc}
\toprule
\multicolumn{6}{c}{\textbf{Rotation noise std. (rad) $\rightarrow$}}\\
Methods & 0 & 0.005 & 0.01 & 0.02 & 0.05 \\
\midrule
MapTRv2 & 61.5 & 60.8 & 58.0 & 50.6 & 32.3 \\
\makecell{MapTRv2 +\\OptiMVMap} & 71.0 & 70.8 & 70.6 & 69.8 & 65.7 \\
\midrule
\multicolumn{6}{c}{\textbf{Translation noise std. (m) $\rightarrow$}}\\
Methods & 0 & 0.05 & 0.1 & 0.5 & 1.0 \\
\midrule
MapTRv2 & 61.5 & 61.2 & 59.8 & 35.6 & 18.8 \\
\makecell{MapTRv2 +\\OptiMVMap} & 71.0 & 70.8 & 70.7 & 68.0 & 62.3 \\
\bottomrule
\end{tabular}
}
\end{table}

\section{Performance on Geospatially Disjoint Split.}
Tab.~\ref{tab:disjoint} shows results on StreamMapNet's disjoint split. OptiMVMap outperforms SOTA methods HRMapNet (+1.2) and MapExpert (+0.5 mAP), confirming generalization beyond training locations.

\begin{table}[h]
\centering
\small
\caption{\textbf{Performance on StreamMapNet Geo-Disjoint Split.}}
\begin{tabular}{l|c|c|c|c|c}
\toprule
Method  &epoch & $AP_{div}$ & $AP_{ped}$ & $AP_{bnd}$ &$mAP$ \\
\midrule
StreamMapNet &24 &30.1 &29.6 &41.9 &33.9\\
HRMapNet &24 &30.3 &36.9 &44.0 &37.1 \\
Ours &24 &\textbf{32.0} &\textbf{38.1} &\textbf{45.0} &\textbf{38.3}\\
\midrule
MapTracker & 72 & 30.0 &45.9 &45.1 &40.3\\
MapExpert &100 &34.1 &46.7 &45.1 &42.0\\
Ours &72 &\textbf{35.2} &\textbf{47.8} &\textbf{45.9} &\textbf{42.5}\\
\bottomrule
\end{tabular}
\label{tab:disjoint}
\end{table}

\section{Computational Efficiency.}
Tab.~\ref{tab:efficiency} shows our method at $K$=1 achieves near real-time inference comparable to MapTRv2. FPS decreases gradually with $K$, remaining acceptable for offline pipelines. OVS operates on low-resolution uncertainty maps, adding only 12.6ms/sample.

\begin{table}[t]
\centering
\caption{Computational Efficiency.}
\label{tab:efficiency}
\begin{tabular}{lccc}
\toprule
\multicolumn{4}{c}{\textbf{Map Construction Efficiency}} \\
\midrule
\textbf{Baseline} & \textbf{FPS} & \textbf{Ours} & \textbf{FPS} \\
\midrule
HRMapNet & 17.0 & $K$=1 & 10.6 \\
StreamMapNet & 14.2 & $K$=3 & 5.4 \\
MapTRv2 & 14.1 & $K$=5 & 4.8 \\
\midrule
\midrule
\multicolumn{4}{c}{\textbf{Vehicle Selection Latency}} \\
\midrule
 & \textbf{Random} & \textbf{Closest} & \textbf{OVS} \\
\midrule
\textbf{Latency (ms)} & 0.27 & 0.28 & 12.6 \\
\bottomrule
\end{tabular}
\end{table}

\section{Extra Qualitative Results}
\begin{figure*}
    \centering
    \includegraphics[width=1.0\textwidth]{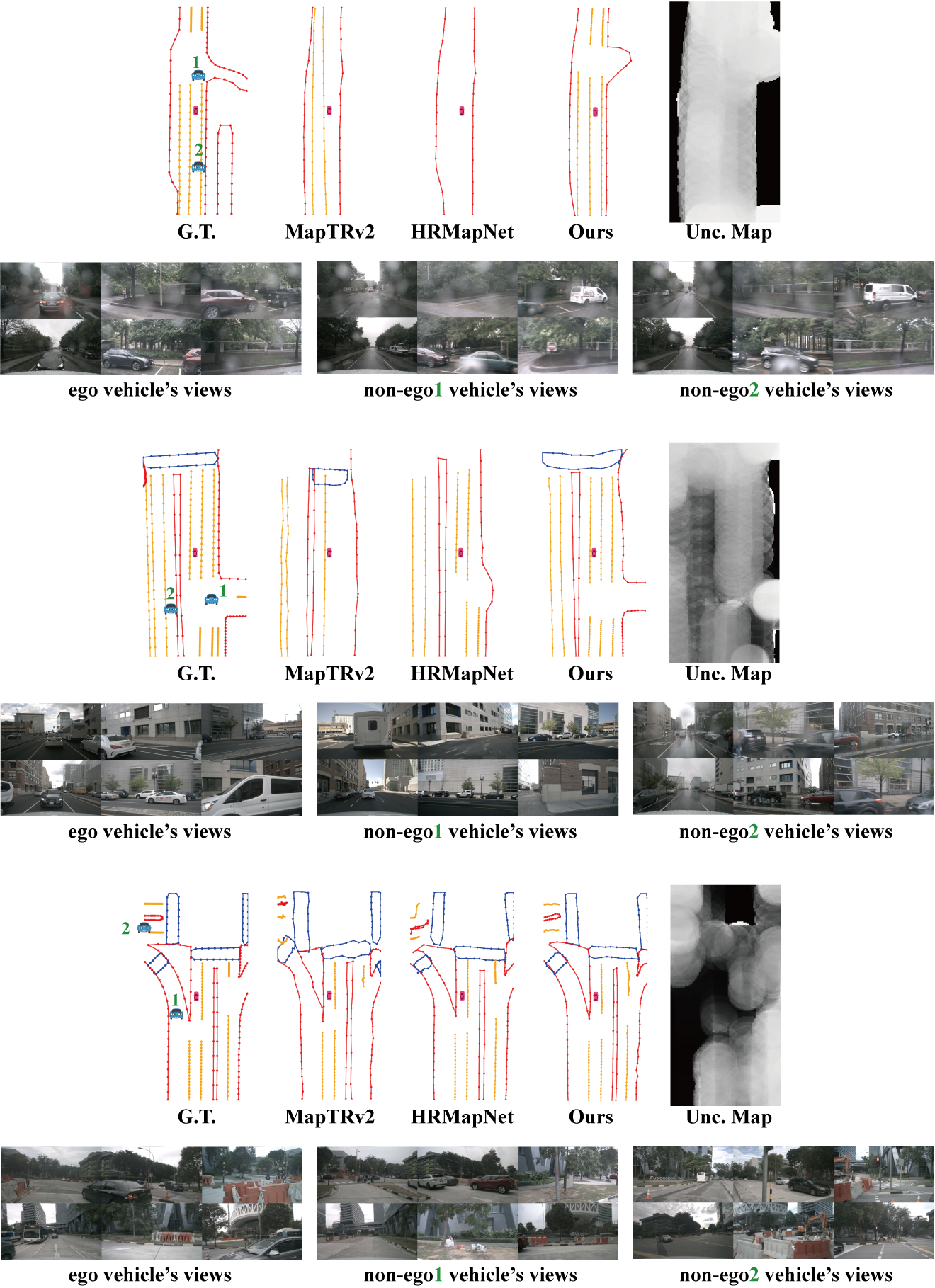}
    \caption{More qualitative results on the nuScenes dataset.}
    \label{fig:vis-res}
\end{figure*}
\begin{figure*}
    \centering
    \includegraphics[width=1.0\textwidth]{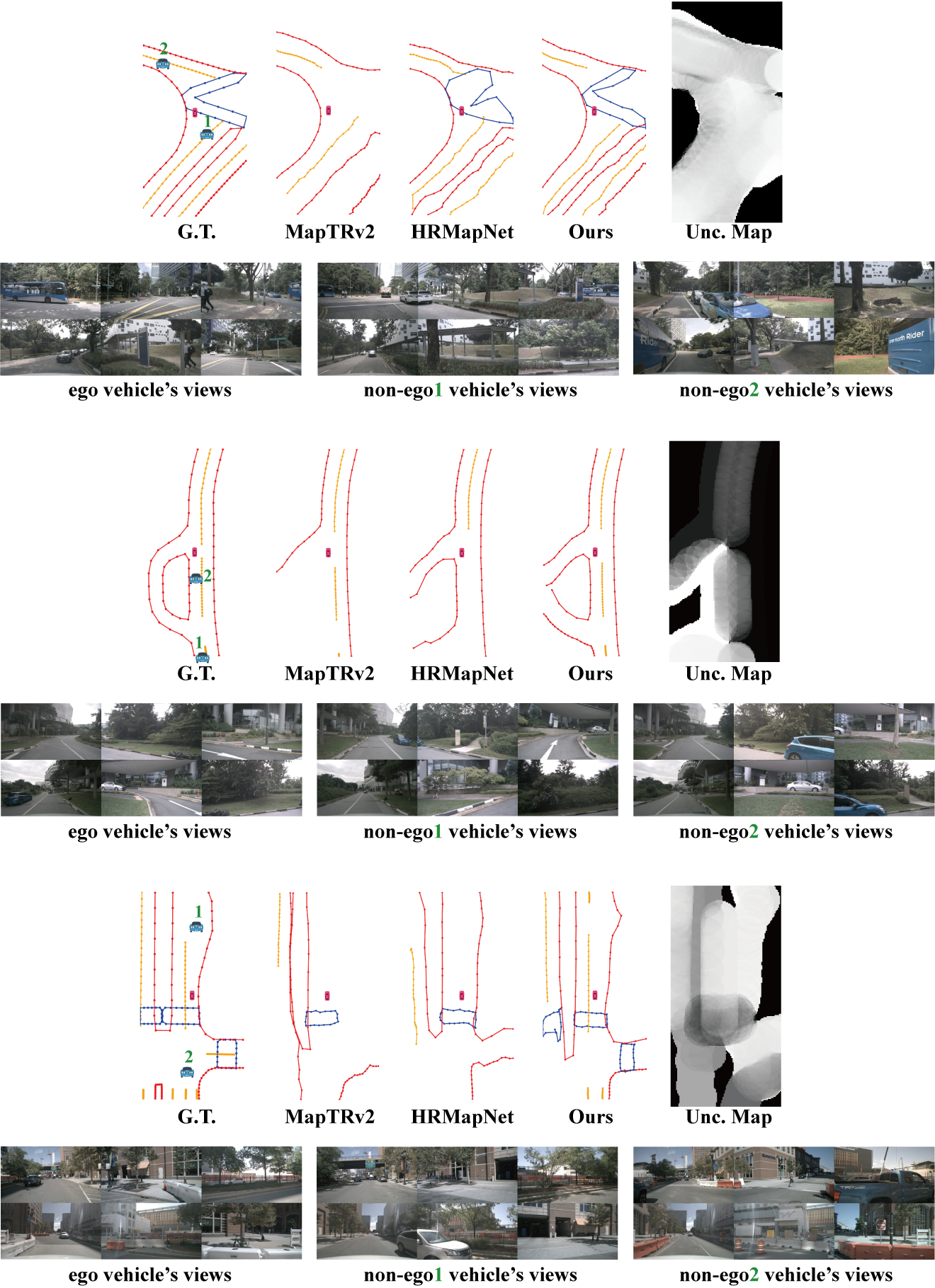}
    \caption{More qualitative results on the nuScenes dataset.}
    \label{fig:vis-res}
\end{figure*}
\begin{figure*}
    \centering
    \includegraphics[width=1.0\textwidth]{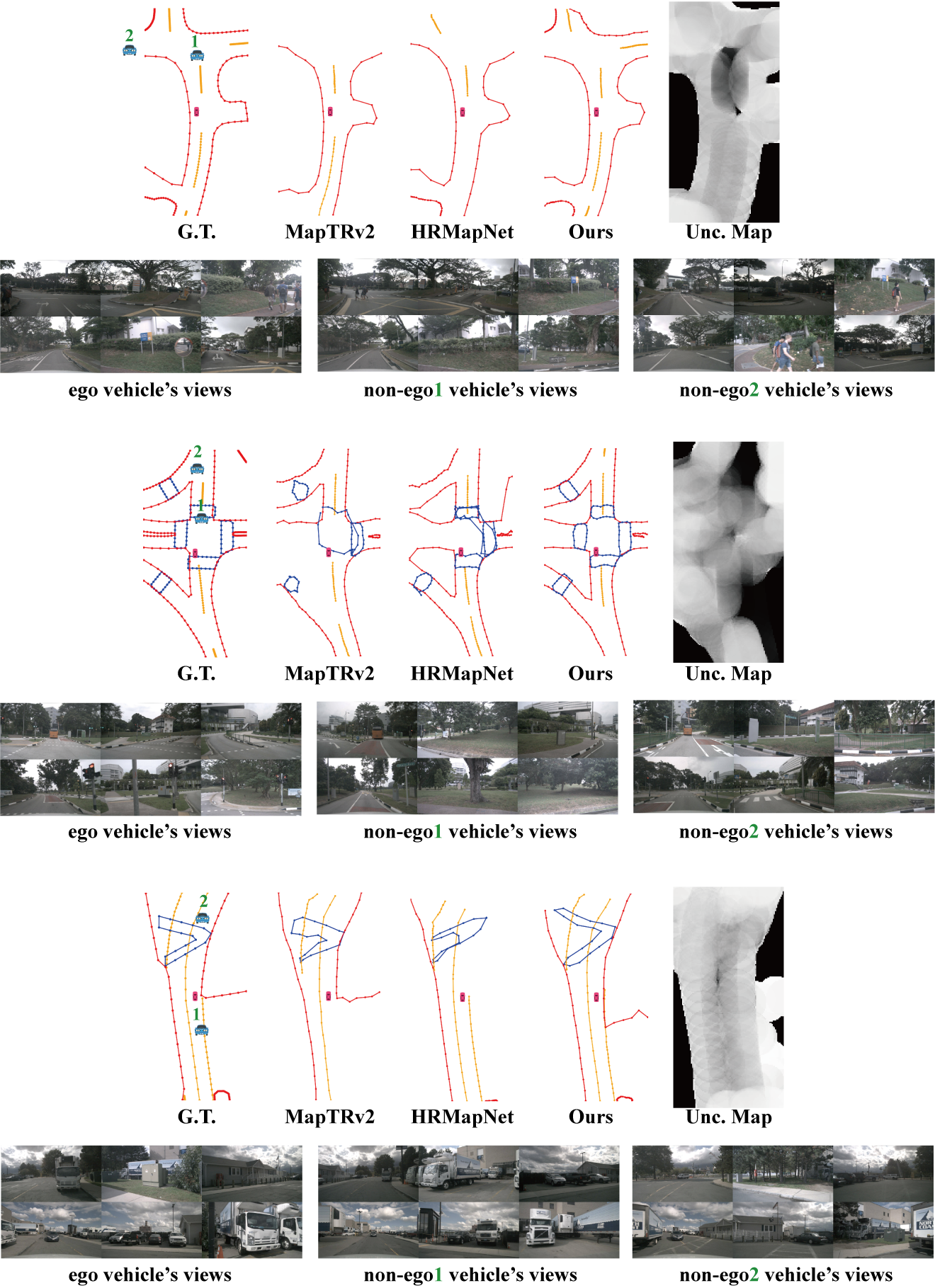}
    \caption{More qualitative results on the nuScenes dataset.}
    \label{fig:vis-res}
\end{figure*}
\begin{figure*}
    \centering
    \includegraphics[width=1.0\textwidth]{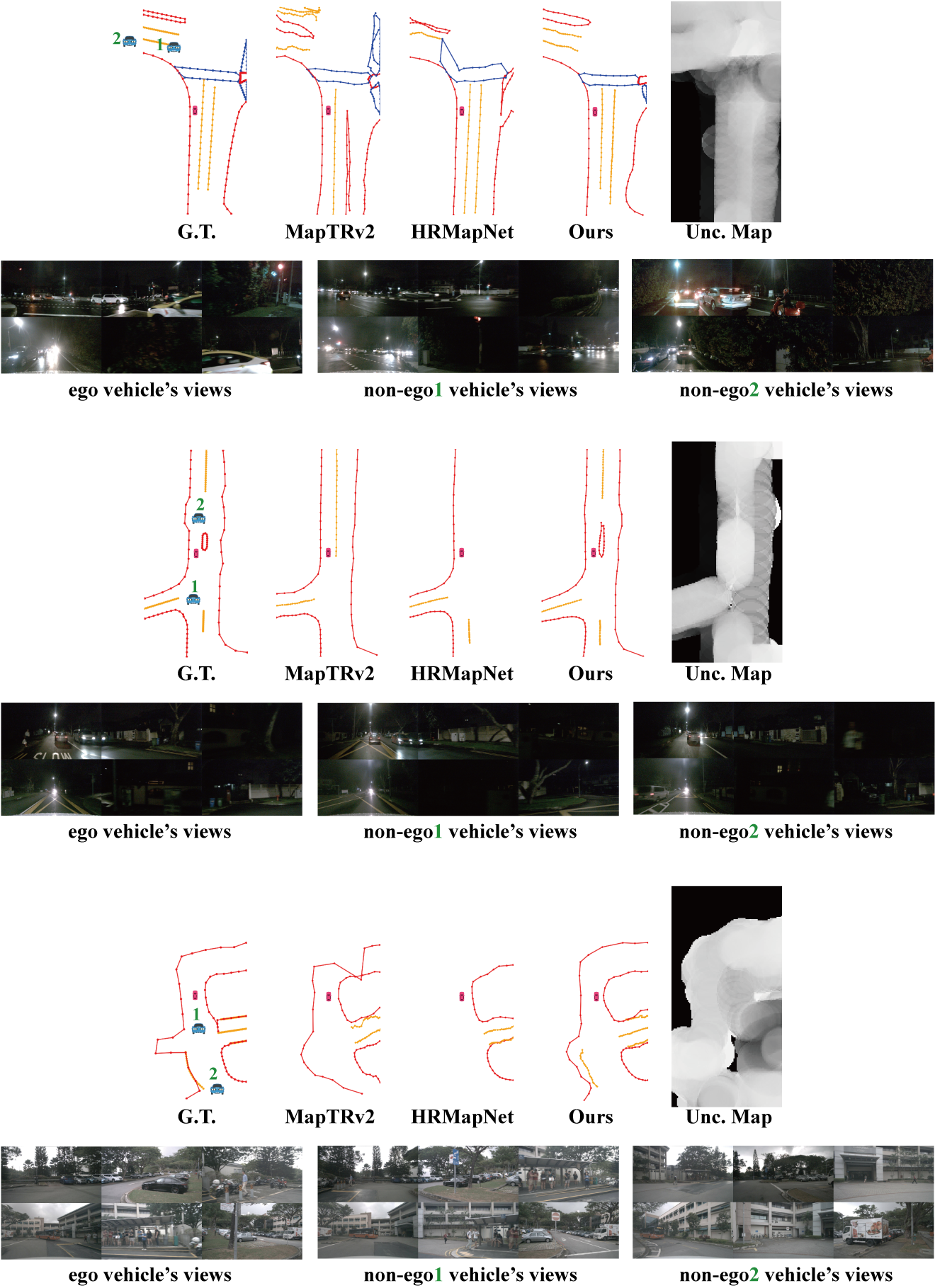}
    \caption{More qualitative results on the nuScenes dataset.}
    \label{fig:vis-res}
\end{figure*}
\begin{figure*}
    \centering
    \includegraphics[width=1.0\textwidth]{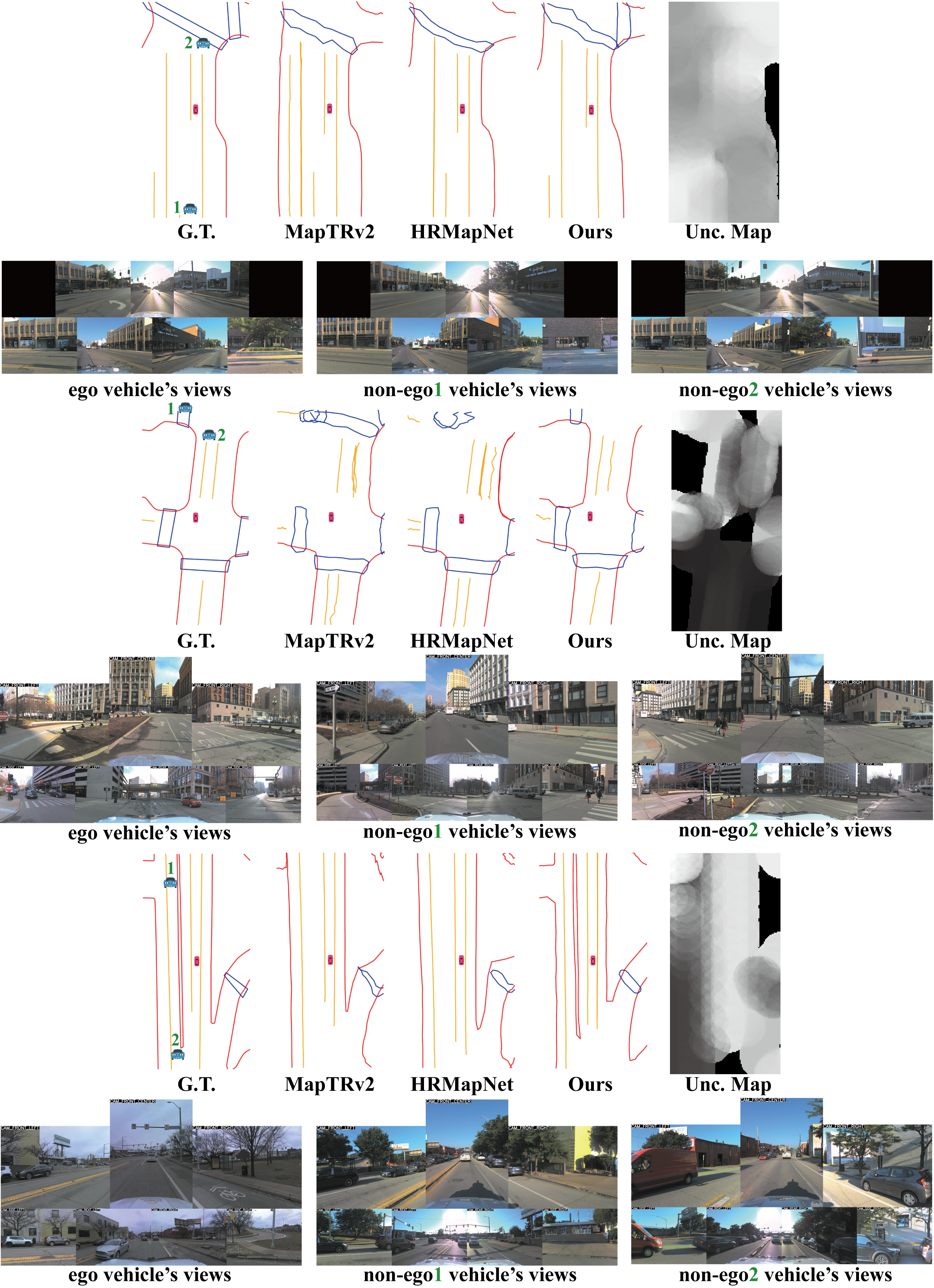}
    \caption{More qualitative results on the Argoverse2 dataset.}
    \label{fig:vis-res}
\end{figure*}
We present 15 additional qualitative examples from the NuScenes and the Argoverse2 dataset using only 3 helper vehicles, covering challenging scenarios such as night-time crossroads, severe occlusion in rainy day, and distant, curved lane-shape. These results reveal that OptiMVMap can select the right non-ego vehicles that provide complementary information for the uncertainty areas while not introducing too much noise. This enables the map decoder to restore missing map details due to occlusion or remote distance, and produce markedly higher-precision reconstructions. By contrast, HRMapNet occasionally underperforms even MapTRv2 in these challenging settings, indicating that the absence of explicit noise filtering and data-source selection in HRMapNet leads to error accumulation and ultimately degrades map quality.